\documentclass[runningheads]{llncs}
\usepackage{amssymb}
\usepackage{amsmath}
\usepackage{graphicx}
\usepackage{tikz}
\usepackage[export]{adjustbox}
\usepackage{xcolor}
\usepackage{graphicx}
\usepackage{tabularx}
\usepackage{booktabs}
\usepackage{dcolumn}
\usepackage{hyperref}
\usepackage{subcaption}

\usepackage[tableposition = top]{caption}

\definecolor{myyellow}{rgb}{1,1,0}
\definecolor{myred}{rgb}{1,0,0}
\definecolor{mygreen}{rgb}{0,1,0}
\definecolor{myblue}{rgb}{0,0,1}

\usepackage{makecell}

\graphicspath{ {./images/} }
\usepackage{etoolbox}

\usepackage{kantlipsum} 

\newcommand{\repeatthanks}{\textsuperscript{\thefootnote}}

\begin{document}
\title{MOBDrone: a Drone Video Dataset for \\ Man OverBoard Rescue
 \thanks{
 Supported by NAUSICAA - “NAUtical Safety by means of Integrated Computer-Assistance Appliances 4.0", a project funded  by the Tuscany region (CUP D44E20003410009).
 } 
}

\titlerunning{MOBDrone}

\author{
Donato Cafarelli \inst{1}\thanks{Co-first authors}\and
Luca Ciampi \inst{1}\repeatthanks \and 
Lucia Vadicamo \inst{1}\repeatthanks \and \\
Claudio Gennaro \inst{1}\and
Andrea Berton \inst{3}\and
Marco Paterni \inst{2}\and\\
Chiara Benvenuti \inst{2} \and 
Mirko Passera\inst{2} \and
Fabrizio Falchi \inst{1}
}
\authorrunning{
D. Cafarelli et al.
}
%
\institute{
Institute of Information Science and Technologies, CNR \\
\and
Institute of Clinical Physiology, CNR 
\and Institute of Geosciences and Earth Resources, CNR \\ Via G. Moruzzi 1, 56124 Pisa, Italy\\
\email{lucia.vadicamo@isti.cnr.it}
}

%
\maketitle              
\begin{abstract}

Modern Unmanned Aerial Vehicles (UAV) equipped with cameras can play an essential role in speeding up the identification and rescue of people who have fallen overboard, i.e., man overboard (MOB). To this end, Artificial Intelligence techniques can be leveraged for the automatic understanding of visual data acquired from drones. However, detecting people at sea in aerial imagery is challenging primarily due to the lack of specialized annotated datasets for training and testing detectors for this task. To fill this gap, we introduce and publicly release the MOBDrone benchmark, a collection of more than 125K drone-view images in a marine environment under several conditions, such as different altitudes, camera shooting angles, and illumination. We manually annotated more than 180K objects, of which about 113K man overboard, precisely localizing them with bounding boxes. Moreover, we conduct a thorough performance analysis of several state-of-the-art object detectors on the MOBDrone data, serving as baselines for further research.




\end{abstract}

\section{Introduction}

The 2021 Annual Overview of Marine Casualties and Incidents \cite{european2021annual} 
reported that $22,532$ marine casualties and incidents were occurred between 2014 and 2020 in the waters of EU Member States or involving EU ships. $7,051$ of these events involved people, with $550$ lives lost and $6,921$ injured. The main events that resulted in fatalities were ship collisions and people slipping/falling into the water. 
Of the falls, 9.8\% were falling overboard, resulting in $84$ lives lost. Survival chances in a Man Overboard (MOB) incident depend on many variables, including the height of the fall, the water temperature, the sea state, and the weather conditions, along with the rescue operation time, the person's state of consciousness and ability to swim, to name but a few. Unfortunately, in most cases (estimated between 85-90\%), it ends in death \cite{garay2017}. 
Indeed, the rescue operations are usually long and complicated. If the person falls overboard while the boat is navigating (e.g., at a speed of 18 knots), the time that elapses from when the alarm is given to when the boat can slow down and turn 180° to return to the MOB point is several minutes. Not to mention that, for safety reasons, the boat cannot turn back rapidly as it would risk running over the victim. Moreover, since the exact rescue point is not always detectable due to sea currents and alarm delays, it is clear that the MOB scenario is very critical and dangerous.

Quick and effective search and rescue operations (SAR) are crucial to increasing the victim's chances of survival. To this end, it is essential to determine a limited search area and plan paths for rescue boats \cite{mou2021cooperative}. 
Unmanned Aerial Vehicles (UAVs) equipped with thermal and/or video cameras can be profitably used to localize and track people overboard, thus expediting rescue operations and increasing their probability of success. In this regard, the ``NAUtical Safety by means of Integrated Computer-Assistance Appliances 4.0" (NAUSICAA) 
project aims at creating a system for medium and large boats in which the conventional control, propulsion, and thrust systems are integrated with a series of latest generation sensors (including lidar systems, cameras, radar, drones) for assistance during the navigation and mooring phases. Specifically,  within the project, we will use commercial aerial drones (equipped with a video camera) and Artificial Intelligence (AI) techniques to search for people overboard automatically.

Many AI techniques have achieved outstanding results in localizing and recognizing people and objects in images and video frames in recent years \cite{fasterrcnn,centernet,yolov3}. 
However, evaluating these approaches (or developing new ones) in a MOB scenario is difficult due to the lack of labeled data. Although many annotated datasets containing people and objects in everyday scenarios are publicly available, to the best of our knowledge, the same cannot be said for the case of aerial footage of people and objects in marine environments. To fill this gap, we collected and publicly released \cite{dataset_zenodo} a large-scale dataset of aerial footage of people who, being in the water, simulated the need to be rescued. Our dataset, named \textit{MOBDrone}, contains 66 video clips with $126,170$ frames manually annotated with more than $180$K bounding boxes (of which more than $113$K belonging to the \textit{person} category). The videos were gathered from one UAV flying at an altitude of 10 to 60 meters above the mean sea level. 

This paper introduces our dataset and describes the data collection and annotation processes. Moreover, it presents an in-depth experimental analysis of the performance of several state-of-the-art object detectors on this newly established MOB scenario, serving as baselines. We hope that this benchmark and the preliminary results may become a reference point for the scientific community concerning the localization of MOBs from UAV imagery.


Evaluation code and all other resources for reproducing the results are available at
\url{http://aimh.isti.cnr.it/dataset/MOBDrone}





\section{Related work}\label{sec:relatedwork}
In the last years, many annotated datasets have been released for supporting the supervised learning of modern detectors based on deep neural networks \cite{Ciampi_2020,Amato_2019,coco,Ciampi_2021}. However, only a few include images or videos taken from UAVs, and most are not focused on the marine environment. This section briefly reviews some of these drone-view datasets suitable for object detection.


VisDrone \cite{zhu2021detection} is the largest object detection and tracking dataset in this category. It consists of 179,264 frames extracted from 263 video clips 
captured by various drone-mounted cameras covering different urban and suburban areas under various weather and lighting condition. Frames are manually annotated with more than 2.6 million bounding boxes localizing targets such as pedestrians, vehicles, and bicycles. Another remarkable dataset is UAVTD \cite{du2018unmanned}, suitable for vehicle detection. It consists of 80K images gathered from a UAV platform in different urban scenarios and contains 2,700 vehicles annotated with bounding boxes.
Another annotated dataset for car detection is the MOR-UAV \cite{mandal2020mor}, which comprises more than 10K drone-view images.
Finally, CAPRK \cite{hsieh2017drone} is a view drone dataset exploited for detecting and counting parked vehicles \cite{Amato_2019_counting}.




Few 
works have been done to date on creating datasets of images taken by drones of people in marine environments. Lygouras et al. \cite{lygouras2019unsupervised} 
addressed the problem of open water human detection by conducting real-time recognition onboard a rescue hexacopter. 
They gathered a swimmers dataset composed of 
images collected from the internet and frames recorded from a drone. In total, the dataset consists of just 4,500 full HD images. 
Recently, Varga et al. \cite{Varga_2022_WACV} released the SeaDroneDataset that contains over 54K annotated frames captured from various altitudes and viewing angles. The dataset mainly contains 
people swimming in open water, and the frames are annotated using six classes: swimmer, floater (swimmer with life jacket), swimmer$^\dagger$ (a person on a boat not wearing a life jacket), floater$^\dagger$ (a person on a boat wearing a life jacket), life jacket, and boat. The main difference with our dataset is that we focus on people at sea without a life jacket (since in the fall into the water from a large ship it is unlikely that the person was previously wearing a life jacket), and we also consider different scenarios of the person's state of consciousness. Nevertheless, the SeaDronesSee dataset is an excellent reference for the task of human detection and tracking in the marine environment that we plan to use in the future, at least for some classes, in conjunction with our MOBDrone for training and testing deep neural networks. Finally, Ferau et al. in \cite{feraru2020towards} faced the problem of assisting SAR operations in MOB incidents using autonomous UAV-based systems. Unlike our work, they aim to locate people in the water by analyzing images recorded with thermal instead of video cameras.
The automatic detection and classification of objects in water from thermal images acquired using UAVs was also explored in \cite{leira2015automatic}.

\section{The MOBDrone Dataset} \label{sec:MOBDroneDataset}
Our \textit{MOBDrone Dataset}, which we publicly released at \cite{dataset_zenodo}, aims to overcome the lack of large public datasets of drone-based imagery for overboard human detection. Its realization required nearly 80 hours of work between data acquisition, post-processing, and annotation, involving, among others, a certified pilot of the Fly\&Sense Service of the CNR of Pisa for UAV flight operations and two professional divers for in-water activities.
In the following, we detail the processes of data collection and curation.

\subsubsection{Data Collection.}
We carried out the drone shooting activities in the Gombo beach 
of the Migliarino, San Rossore, and Massaciuccoli Park (Pisa, Italy).
This choice was dictated for privacy reasons and to ensure compliance with the safety protocols of UAV flight operations. Indeed, the Gombo beach is a segregated area that can be accessed only after obtaining the appropriate authorization from the Park Authority.

To guarantee variability of data, we identified several dimensions of interest, including (i) \textit{subjects/objects to be filmed} (people, lifebuoys, boats, rocks, pieces of wood, parts of the land, and whatever else there is naturally), (ii) \textit{person's state of consciousness} (conscious, semiconscious or unconscious), (iii) \textit{person's visual appearance} (man, woman, persons in light, dark or colored clothing, persons in a bathing suit, etc.), (iv) \textit{light changes} (different shooting times), (v) \textit{altitude and camera directions} (e.g., try to fly at high altitudes to see a more significant portion of the sea, and at lower altitudes better to see the objects and a possible man overboard, also changing the camera shooting angle). 

We gathered a total of 49 videos at high resolution (4K) exploiting the DJI FC6310 camera of the Phantom 4 Pro V2 drone. The camera angle was perpendicular to the water (90°), except for a small set of shots where a 45° angle was used. Two professional divers (one male and one female) 
simulated various scenarios of a person overboard, including a conscious person (swimming, floating, or waving their arms to attract attention) and an unconscious person (floating body in a supine or prone position, or partially floating, i.e., part of the body is below the water surface). 
Some videos incidentally captured people close to the portion of the sea where our divers were positioned.
We split these videos into multiple video clips to remove portions where people were identifiable for privacy concerns. 
The final dataset contains 66 videos that we post-processed, as described in the following section.

\subsubsection{Data Curation.}
\begin{table}[htbp] 
\caption{\textbf{Dataset details.} 
The MOBDrone benchmark contains 
126,170 drone-view images at six different heights in MOB 
scenarios.}
\medskip
\centering
 \centering
    \newcolumntype{C}{>{\centering\arraybackslash}X}
    \begin{tabularx}{0.6\textwidth}{crCC}
    \toprule
\textbf{Altitude}& & \textbf{\# Images} &  \textbf{\# Video Clips} \\ 
\midrule
 	 10 m &	&	$958$		&   1		\\
 	 20 m&	&	$10,053$	&	6		\\
 	 30 m&	&	$29,404$	&	15		\\
 	 40 m&	&	$33,046$	&	13		\\
 	 50 m&	&	$29,183$	&	16		\\
 	 60 m&	&	$23,526$	&	15		\\
 	 \midrule
 	 &\textit{tot} &\textbf{126,170} & \textbf{66} \\
\bottomrule
 \end{tabularx}
\label{tab:dataset_details_1}
\end{table}
\begin{figure}[htbp] 
\centering%
{\includegraphics[height=3.3cm, trim=1cm 1cm 0.5cm 0.5cm, clip]{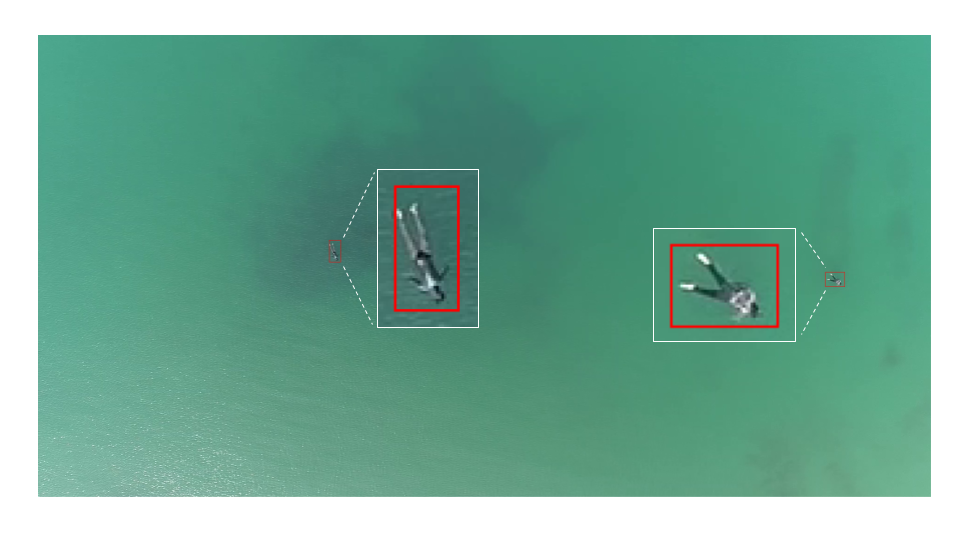}}\hfill
{\includegraphics[height=3.3cm, trim=1cm 1cm 0.5cm 0.5cm, clip]{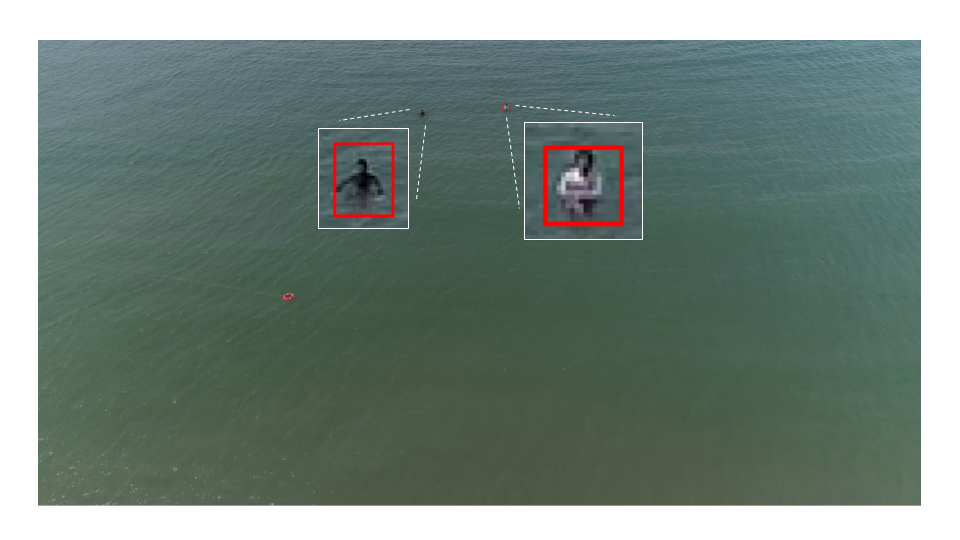}}\\%
\includegraphics[height=3.3cm,, trim=1cm 1cm 0.5cm 0.5cm, clip]{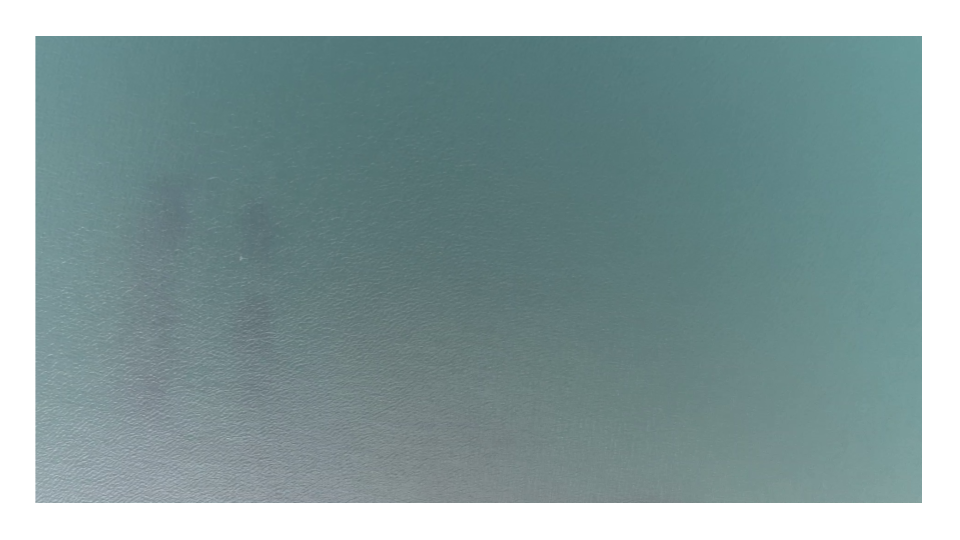}\hfill%
\includegraphics[height=3.3cm, trim=1cm 1cm 0.5cm 0.5cm, clip]{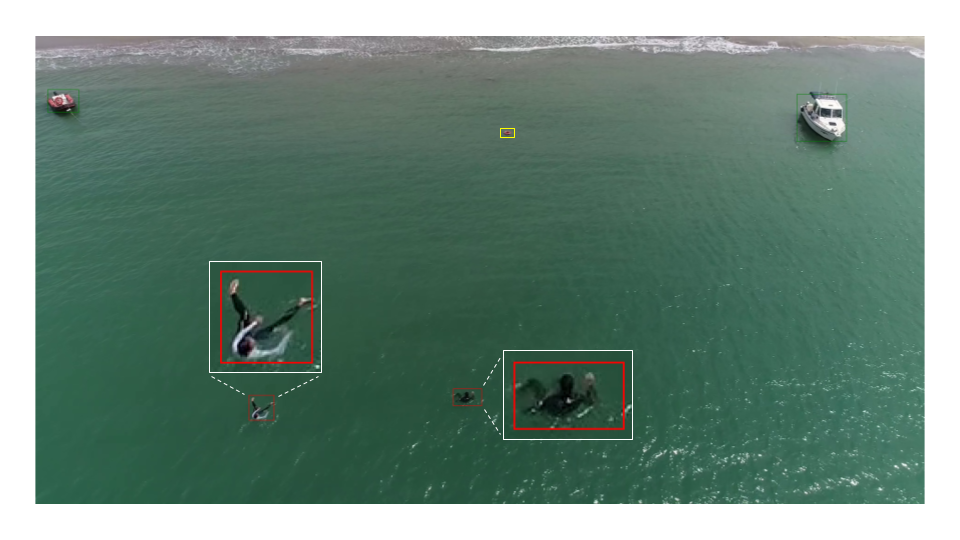}\\%
\includegraphics[height=3.3cm, trim=1cm 1cm 0.5cm 0.5cm, clip]{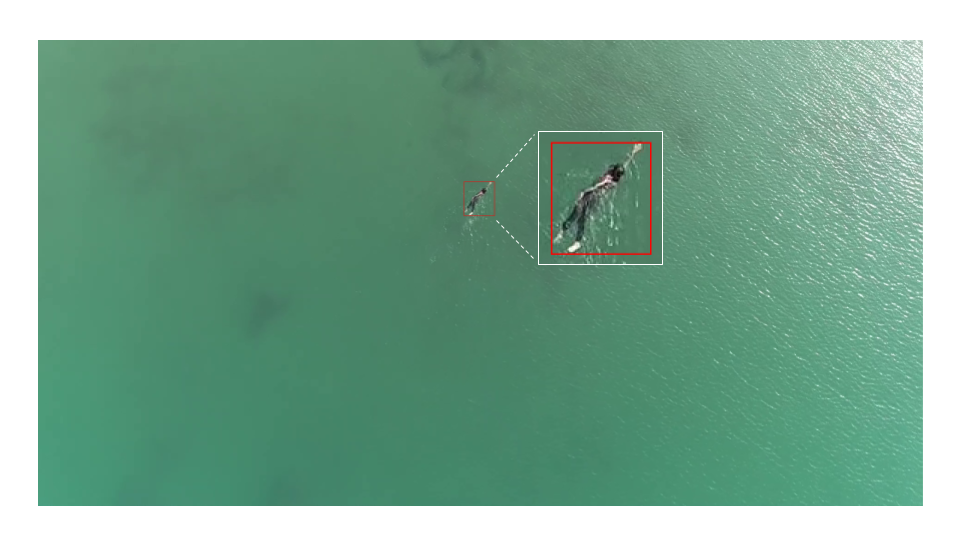}\hfill%
\includegraphics[height=3.3cm, trim=1cm 1cm 0.5cm 0.5cm, clip]{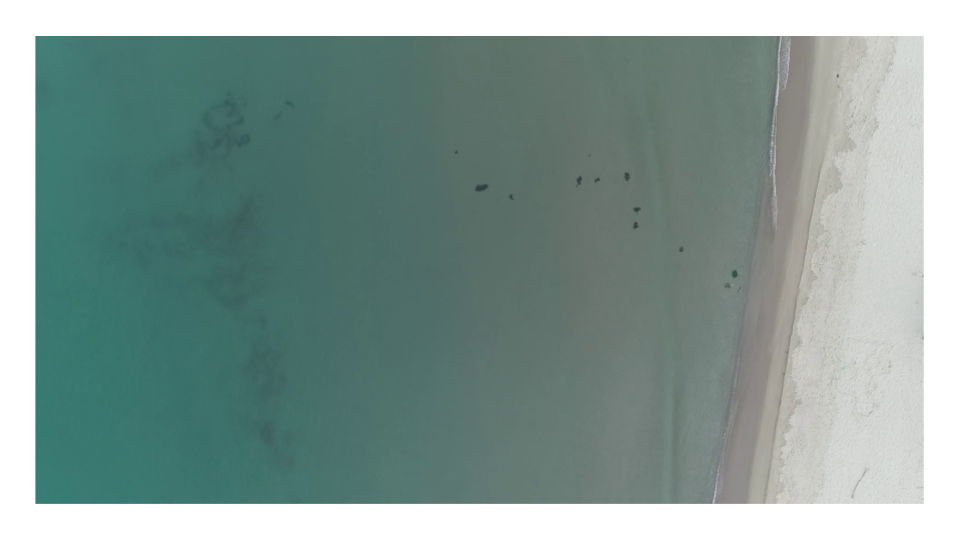}\\%
\caption{\textbf{Samples of the MOBDrone Dataset.} Examples of images captured at different altitudes, light conditions, and camera directions. The bounding box annotations localizing the labeled objects are also shown. Objects belonging to the \textit{person} category, which is the one of paramount interest in MOB scenarios, are outlined with red bounding boxes and zoomed. Note that $27.72\%$ of the images do not contain objects (i.e., images of clear water) and that interfering objects in the background, such as rocks, often trigger false positive detections.}
\label{fig:sample_gt}
\end{figure}
\begin{table}[htbp]
\caption{
\textbf{Annotation statistics.} 
We labeled with bounding boxes 181,689 objects belonging to 5 categories.
}
\medskip
\centering
\newcolumntype{C}{>{\centering\arraybackslash} m{3.2cm}}
\newcolumntype{B}{>{\centering\arraybackslash} m{2.3cm}}
\begin{tabularx}{0.88\textwidth}{BBBC}
\toprule
\textbf{Class} & \textbf{\#Annotations}  & \textbf{\#Images} & \textbf{Samples}\\
\midrule
\textit{person}	&    $113,408$ & $77,365$ & \includegraphics[height=0.75cm, width=3cm]{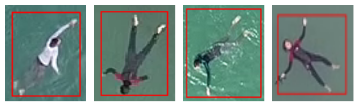}\\ 
\textit{boat}	&    $39,967$ & $31,238$ & \includegraphics[height=0.75cm, width=3cm]{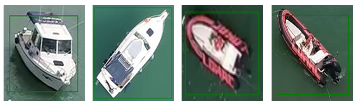}\\ 
\textit{wood}    & 	  $15,980$ & $9,040$ & \includegraphics[height=0.75cm, width=3cm]{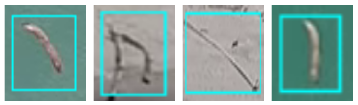}\\
\textit{life buoy }&	$10,401$ & $10,386$ & \includegraphics[height=0.75cm, width=3cm]{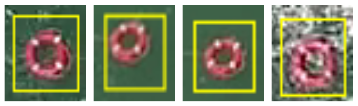}\\
\textit{surfboard} & $1,933$& $1,933$ & \includegraphics[height=0.75cm, width=3cm]{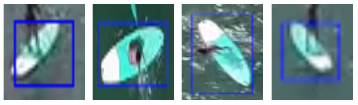}\\
\texttt{no object} && $34,976$ & \includegraphics[height=0.75cm, width=3cm]{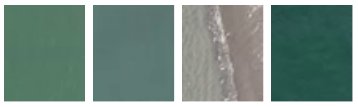}\\
\midrule
\textit{total} & 181,689 &  & \\
\bottomrule
\end{tabularx}
\label{tab:dataset_details_2}
\end{table}

First, we converted the 66 video clips captured in the data acquisition campaign from 4K to 1080p resolution. Then, we extracted the frames from the videos at a rate of 30 FPS, obtaining a total of 126,170 images (see  Table \ref{tab:dataset_details_1} for summary statistics). 
Finally, a human expert annotator manually annotated them. Specifically, the annotation process took approximately 60 hours, and the Computer Vision Annotation Tool (CVAT) \cite{CVAT} 
was used. Although our work focuses on localizing and recognizing people, we also annotated other objects present in the scenes. In particular, we considered a total of 5 classes (\textit{person}, \textit{boat}, \textit{surfboard}, \textit{wood}, \textit{life\_buoy}). We provide a bounding box precisely localizing each instance of the objects of interest. The total number of annotations is $181,689$, of which the ones related to the \textit{person} class, which is of primary interest in the MOB scenario, is $113,408$. However, note that about $27.72\%$ of the images do not contain any objects (i.e., images of clear water).
We report some statistics concerning the annotations in Table \ref{tab:dataset_details_2}, while we show some samples of our dataset in Figure \ref{fig:sample_gt}.

\section{Detection Performance Analysis}
\label{sec:experiments}

In this section, we evaluate several state-of-the-art object detectors on our MOBDrone dataset\footnote{Although in this work we exploited the whole dataset as a test benchmark, in \cite{dataset_zenodo} we provide training and test splits.}, focusing on the detection of the overboard people, i.e., on the localization of the object instances belonging to the class \textit{person}. 
In the first part of our performance analysis, we compare 9 of the most popular and performing object detectors present in the literature. Then, we look upon the best three ones, performing a more in-depth analysis of the obtained results.   


The detection methods considered in our analysis can be roughly grouped into three categories, i.e., anchor-based Convolutional Neural Network (CNN) methods, anchor-free CNN methods, and Transformer-based methods.  We briefly summarize them below. We refer the reader to the papers describing the specific detectors for more details.

\noindent \textit{Anchor-based CNN methods} compute bounding box locations and class labels of object instances exploiting CNN-based architectures that rely on anchors, i.e., prior bounding boxes with various scales and aspect ratios. They can be divided into two groups: i) the two-stage approach, where a first module is responsible for generating  a sparse set of object proposals and a second module is in charge of refining these predictions and classifying the objects; and ii) the one-stage approach that directly regresses to bounding boxes by sampling over regular and dense locations, skipping the region proposal stage. Here, we use Faster R-CNN \cite{fasterrcnn} and Mask R-CNN \cite{mask_rcnn} regarding the first group, and YOLOv3 \cite{yolov3}, TOOD \cite{tood} and VarifocalNet (VfNet) \cite{vfnet} concerning the second one. 

\noindent \textit{Anchor-free CNN methods} rely on the prediction of key-points, such as corner or center points, to predict the objects, instead of using anchor boxes and their inherent limitations. In this work, we exploit CenterNet \cite{centernet} and YOLOX \cite{yolox}.

\noindent \textit{Transformer-based methods} rely on the recently introduced Transformer attention modules in processing image feature maps, removing the need for hand-designed components like a non-maximum suppression procedure or anchor generation. In this paper, we consider DEtection TRansformer (DETR) \cite{detr} and one of its evolution, Deformable DETR \cite{deformable_detr}.
\begin{table}[tbp] 
    \caption{\textbf{Comparison of the considered detectors.} mAP@[0.50:0.95] is the AP averaged over 10 IoU thresholds in the range $[0.50 : 0.95]$ with a step size of $0.05$, while AP50 is the AP computed at the single IoU threshold value of $0.50$.}
    \medskip
    \centering
    \newcolumntype{C}{>{\hsize=0.9\hsize\centering\arraybackslash}X}
    \newcolumntype{B}{>{\hsize=1.2\hsize}X}
    \begin{tabularx}{\textwidth}{BCC}
    \toprule
     Method & AP50 $\uparrow$ & mAP@[0.50:0.95] $\uparrow$ \\
    \midrule
     VarifocalNet \cite{vfnet}  & \textbf{0.378} & \textbf{0.144}  \\  
     TOOD \cite{tood} & 0.314 & 0.116 \\ 
     Deformable DETR \cite{deformable_detr} & 0.199 & 0.075 \\   
     YOLOX \cite{yolox}  & 0.126 & 0.049  \\
     Faster R-CNN \cite{fasterrcnn} & 0.126 & 0.041 \\
     CenterNet \cite{centernet}  & 0.124 & 0.041  \\   
     DETR \cite{detr} & 0.128 & 0.040 \\   
     Mask R-CNN \cite{mask_rcnn} & 0.109 & 0.033 \\   
     YOLOv3 \cite{yolov3}  & 0.011 & 0.009  \\
    \bottomrule
    \end{tabularx}
    \label{tab:map_all_models}
\end{table}

We evaluate and compare the above-described detectors over our MOBDrone dataset following the golden standard Average Precision (AP), i.e., the average precision value for recall values over 0 to 1. Specifically, we consider the MS COCO mAP@[0.50:0.95] \cite{coco}, i.e., the AP averaged over 10 IoU thresholds in the range $[0.50 , 0.95]$ with a step size of $0.05$, and the AP50, i.e., the AP computed at the single IoU threshold value of $0.50$. We refer the reader to \cite{coco} for more details. All the detection techniques that we employed were pre-trained\footnote{Pre-trained models are available, e.g., in the model zoo of MMDetection project \cite{mmdetection}} on the COCO dataset \cite{coco}, a popular collection of images in everyday contexts comprising objects belonging to 80 different categories, of which is present the \textit{person} class. To evaluate the performance of the detectors, we filtered the obtained detections considering only the ones classified as \textit{person}. We report the obtained results in Table \ref{tab:map_all_models}. 
The model which turns out to be the most performing is VarifocalNet, considering both the metrics, followed by TOOD and Deformable DETR. However, in general, all the detectors exhibit moderate performance, indicating the difficulties in localizing persons in this challenging scenario. We deem that the most significant metrics in our scenario is the AP50, since i) the dataset is manually labeled by humans and therefore is accurate in terms of classification and inaccurate in terms of boundaries, ii) it is \textit{not} crucial to precisely localize instances, i.e., it is critical to detect overboard persons but the quality of the localization is less important. With this in mind, in the following, we show an in-depth analysis of the three AP50 best models, i.e., VarifocalNet, TOOD, and Deformable DETR.


\begin{figure}[tbp]
   \centering
   \subfloat[Precision vs. Recall curves (IoU=0.5). Areas under curves correspond to AP50.
   ]
   {\includegraphics[height=5.1cm,width=6cm]{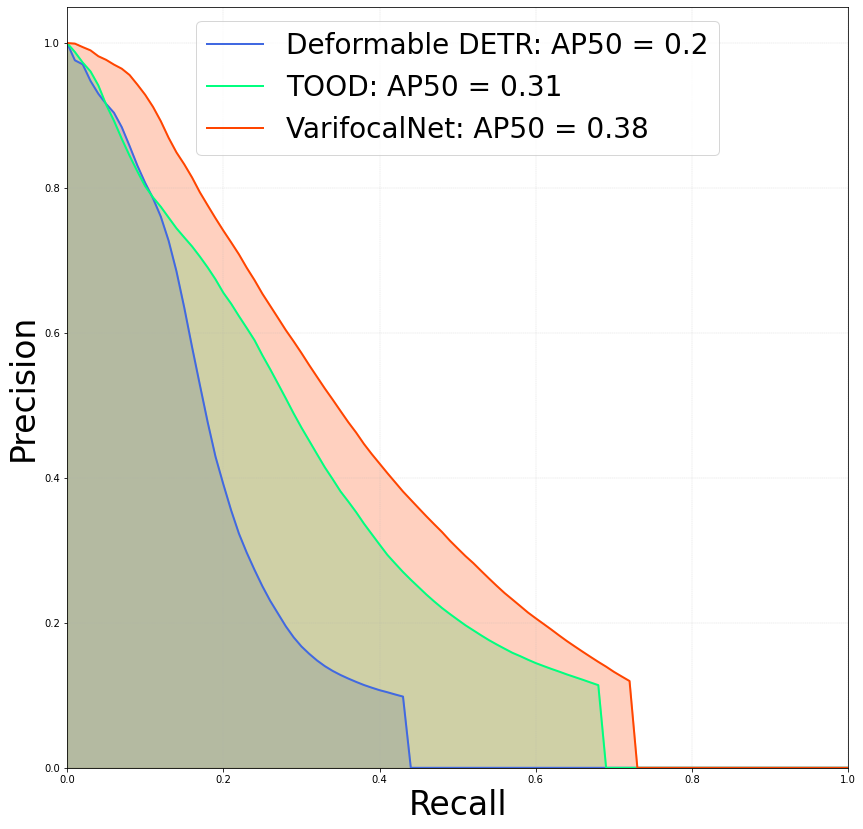}\label{fig:comparison_pr_curve_3_best_models}}
   \hfill
   \subfloat[ $F_1$-score vs. detection threshold curves (IoU=0.5).
   ]{\includegraphics[height=5.1cm,width=6cm]{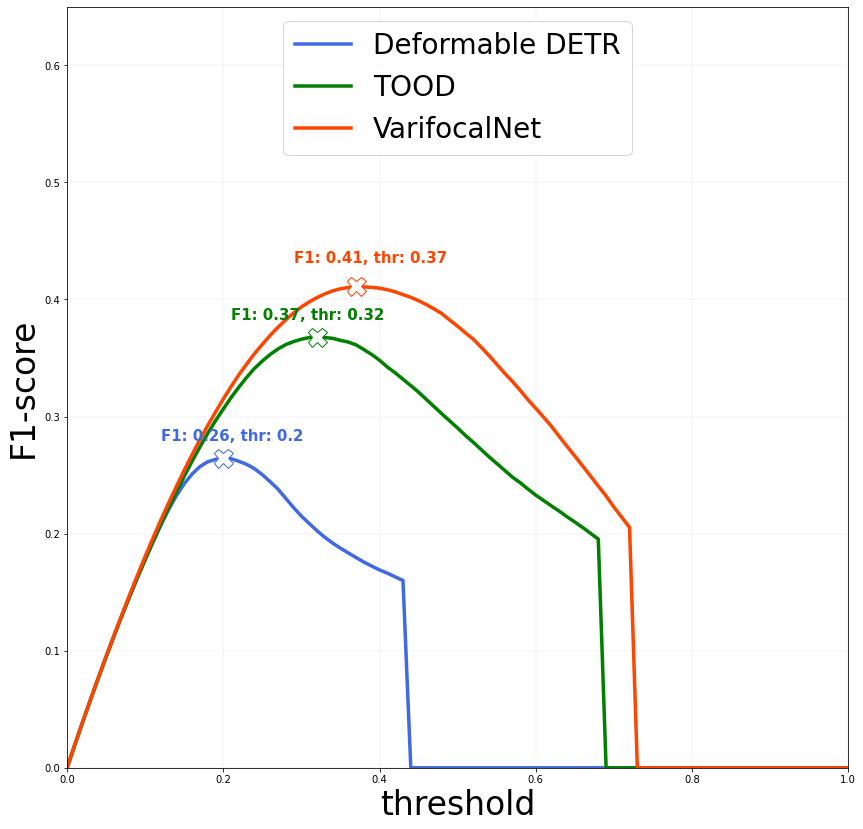}\label{fig:comparison_f1-thr_curve_3_best_models}}
   \caption{\textbf{Comparison of the three best detectors}. We report Precision-Recall (a), and $F_1$-detection threshold (b) curves of the three best models (VfNet, TOOD, and Deformable DETR). VfNet shows best performances.}
   \label{fig:comparison_best_3_models}
 \end{figure}
 \begin{figure}[tp] 
\centering%
{\includegraphics[height=2.8cm, trim=1cm 1cm 0.5cm 0.5cm, clip]{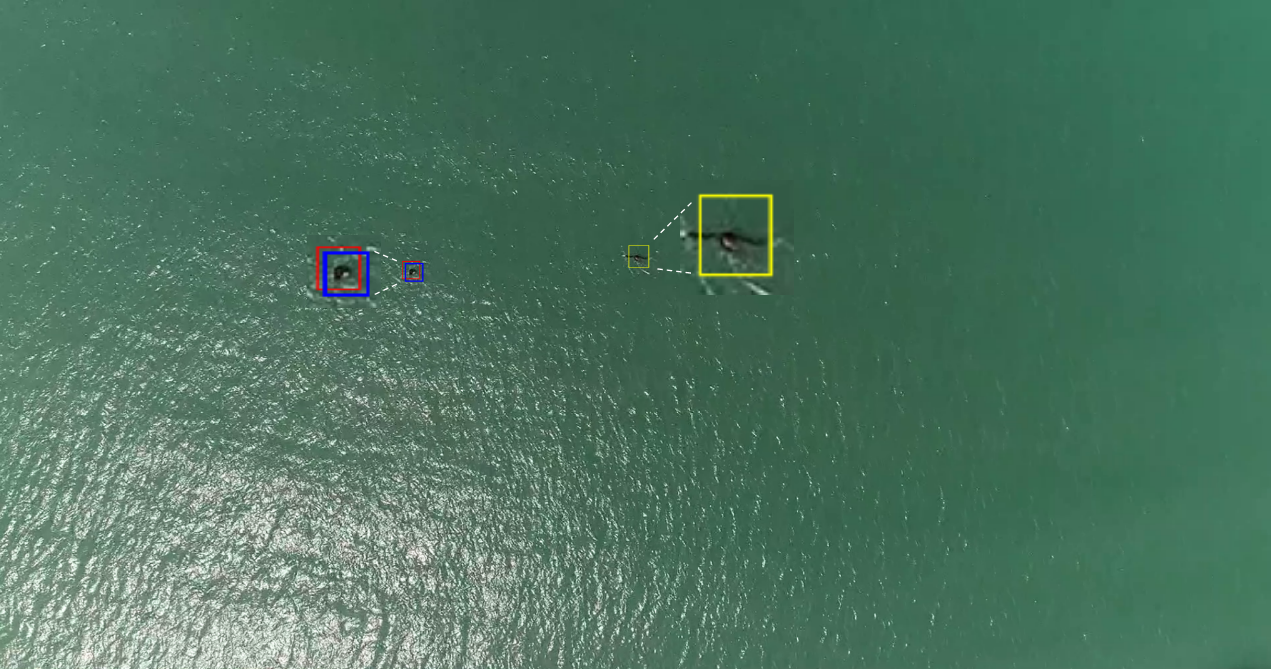}}\hfill
{\includegraphics[height=2.8cm, trim=1cm 1cm 0.5cm 0.5cm, clip]{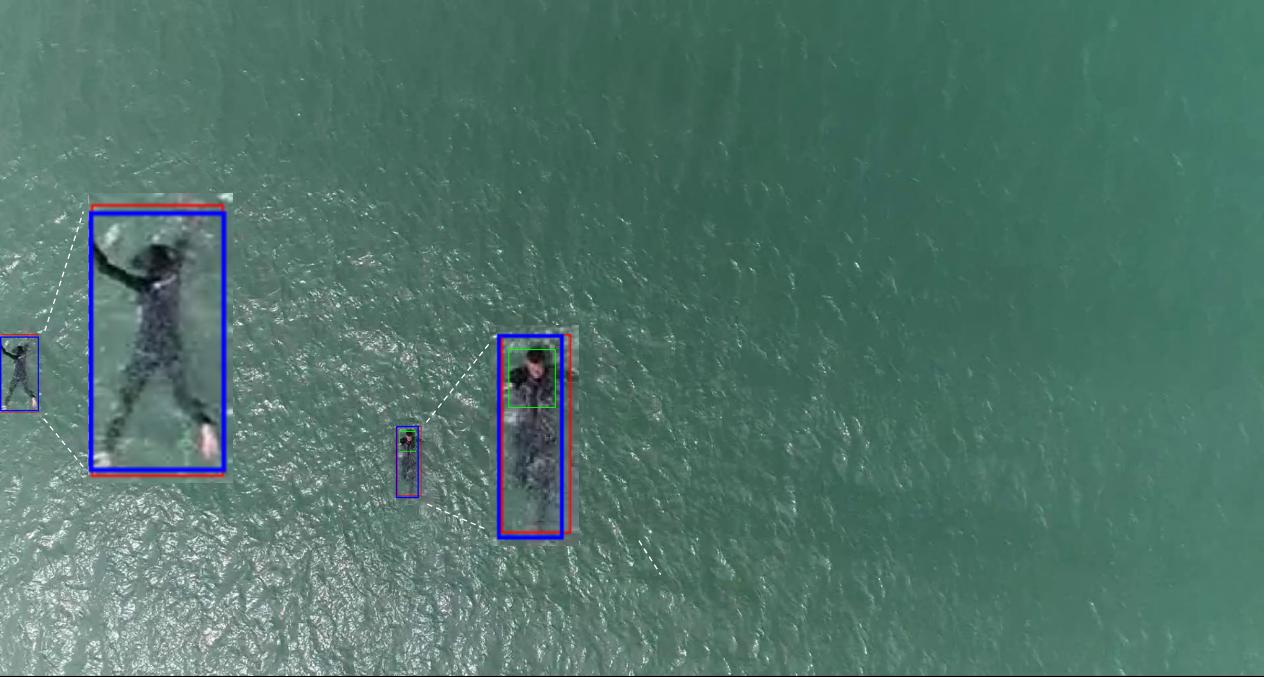}}\\%
\caption{\textbf{Detections produced by VarifocalNet.} We indicate \colorbox{mygreen}{false positives} in green, \colorbox{myyellow}{false negatives} in yellow, \colorbox{myblue}{true positive} in blue and \colorbox{myred}{gt} in red.}
\label{fig:sample_det}
\end{figure}
In Figure \ref{fig:comparison_pr_curve_3_best_models}, we report the Precision-Recall curves, i.e.,  precision and recall values for different detection confidence thresholds, of these three best detectors while setting the IoU threshold at 0.50. Areas under these curves correspond to AP50 values. As can be seen, the VarifocalNet detector exhibits the best performance at all confidence thresholds. The same trend is confirmed in Figure \ref{fig:comparison_f1-thr_curve_3_best_models}, where we show $F_1$-score values (where $F_{1} = 2 \times \frac{Precision \times Recall}{Precision + Recall}$) at different detection confidence thresholds, again setting the IoU threshold at 0.50. Still, VarifocalNet shows superior performance compared to the other two detectors. Please note that the maximum values of these curves indicate  
the detection confidence score that may be used by potential users, enclosing a trade-off between the resulting Precision and Recall values. Figure \ref{fig:sample_det} shows some qualitative outputs produced by VarifocalNet using this confidence score.

In Table \ref{tab:res_heights_best_models}, we show a comparison of the best three detectors at different altitudes in terms of AP50 and $F_1$-score. As expected, in general, performances decrease with increasing altitude. However, it is interesting to note that TOOD and Deformable DETR particularly struggle to detect small objects, i.e., when the altitude is above 40 meters, while achieving comparable or even better results than VarifocalNet at altitudes below 30 meters.

\begin{table}[tbp]
    \centering
    \caption{\textbf{Comparison of the three best detectors at different altitudes.} 
    AP50 and $F_1$ are the AP and the $F_1$-score computed with IoU set to $0.50$.}
    \medskip
    \label{tab:res_heights_best_models}
    \newcolumntype{C}{>{\centering\arraybackslash}X}
    \newcommand*\mc[1]{\multicolumn{2}{c}{{#1}}}
    
    \begin{tabularx}{\textwidth}{CCCCCCC}
    \toprule
    & \mc{VarifocalNet \cite{vfnet}} & \mc{TOOD \cite{tood}} & \mc{{Deformable DETR \cite{deformable_detr}}} \\
    \cmidrule(lr){2-3} \cmidrule(lr){4-5} \cmidrule(lr){6-7} 
    Altitude & AP50 $\uparrow$ & $F_1$ $\uparrow$ & AP50 $\uparrow$ & $F_1$ $\uparrow$ & AP50 $\uparrow$ & $F_1$ $\uparrow$ \\
    \midrule
     10 m  & 0.973  & 0.444 & \textbf{0.989} & 0.363 & 0.959 & \textbf{0.636}  \\
     20 m  & \textbf{0.771}  & \textbf{0.318} & 0.681 & 0.308 & 0.514 & 0.279 \\
     30 m  & 0.400  & 0.199 & \textbf{0.407} & \textbf{0.223} & 0.240 & 0.210  \\
     40 m  & \textbf{0.540}  & \textbf{0.226} & 0.406 & 0.203 & 0.314 & 0.209  \\
     50 m  & \textbf{0.241}  & \textbf{0.161} & 0.187 & 0.140 & 0.107 & 0.08  \\
     60 m  & \textbf{0.205}  & \textbf{0.223} & 0.171 & 0.196 & 0.063 & 0.131  \\
    \bottomrule
    \end{tabularx}
\end{table}

    

\begin{table}[tbhp]
    \centering
    \caption{\textbf{Classwise Analysis.} 
      We consider the detections of
      all 80 COCO classes, accounting for errors due to misclassified objects, i.e., detected objects that matched with \textit{person} annotations but that were classified as objects of another category. 
     TPR is the \textit{True Positive Rate} with respect to the  target \textit{person} class.
    dTPR 
    is the ratio of person instances correctly detected but misclassified. 
    The overall \textit{detection Recall} (dR) is the proportion of detected person instances considering also misclassified objets; the overall \textit{detection Miss Rate} is the proportion of person instances that were not detected at all.
    We set IoU to $0.5$.
    }
    \medskip
    \label{tab:class_analysis_best_models}
    \newcolumntype{C}{>{\hsize=0.91\hsize\centering\arraybackslash}X}
    \newcolumntype{B}{>{\hsize=1.6\hsize}X}
    \newcommand*\mc[1]{\multicolumn{2}{c}{{#1}}}
    \begin{tabularx}{\textwidth}{BCCCCCCC}
    \toprule
    & person & bird & airplane & kite & other & \mc{Overall} \\
    \cmidrule(lr){2-2} \cmidrule(lr){3-3} \cmidrule(lr){4-4} \cmidrule(lr){5-5} \cmidrule(lr){6-6} \cmidrule(lr){7-8}
    Method & TPR $\uparrow$ & dTPR $\uparrow$ & dTPR $\uparrow$ & dTPR $\uparrow$ & dTPR $\uparrow$ & dR $\uparrow$ & dMR $\downarrow$ \\
    \midrule
    VfNet \cite{vfnet} & 0.285 & 0.266 & 0.190 & 0.067 & 0.012 & \textbf{0.818} & \textbf{0.182} \\
    TOOD \cite{tood} & \textbf{0.326} & 0.118 & 0.212 & 0.125 & 0.017 & 0.799 & 0.201 \\
    Def. DETR \cite{deformable_detr} & 0.206 & 0.311 & 0.072 & 0.041 & 0.026 & 0.657 & 0.343 \\
    \bottomrule
    \end{tabularx}
\end{table}

Finally, in Table \ref{tab:class_analysis_best_models}, we report a classwise analysis of the obtained detections, i.e., we consider the detections belonging to all the 80 classes and not only the detections classified as \textit{person}. Specifically, we take into account also errors due to misclassified objects, i.e., detected objects that matched with \textit{person} annotations but that were classified as objects belonging to another category. We define the \textit{True Positive Rate} (TPR) as the ratio between the number of correctly detected and classified person instances ({\small TP}) and the total number of person instances in the ground-truth ({\small P}). On the other hand, we define $\text{dTPR}(c)= \frac{\text{dTP(c)}}{\text{P}}$ as the \textit{detection True Positive Rate} for the output class $c$ with respect to the target \textit{person} class, that is the the number {\small $\text{dTP(c)}$} of person instances detected correctly (i.e., considering only the IoU of predicted and target bounding boxes) but classified with category $c$ divided by the total number of person instances in the ground-truth. In other words, $\text{dTPR}(c)$ gives us the proportion of person instances that were detected correctly but misclassified with category $c$. The sum of the TPR and all the $\text{dTPR}(c)$ gives the overall \textit{detection Recall} (dR), i.e., the ratio of person instances detected correctly without considering the output classification. Similarly, the overall \textit{detection Miss Rate}, defined as dMR=1-dR, is the portion of person instances that were not detected at all. For example, from Table \ref{tab:class_analysis_best_models}, we can observe that the pre-trained VarifocalNet correctly detected $81.8\%$ of the ground-thruth person instances even if, in most cases, it misclassified them. This may suggest that the same model fine-tuned on MOB data may have room for growth in localizing  \textit{person} instances.

\section{Conclusion and Future Directions} \label{sec:conclusions}
This paper presents the MOBDrone benchmark, a large-scale drone-view dataset suitable for detecting persons overboard. It is part of the NAUSICAA project aiming at creating a control system that, for the first time, uses aerial and marine drones and augmented and virtual reality to provide increased safety to medium and large vessels. Specifically, we collected more than 125K images, and we manually annotated 
more than 180K objects in a marine environment under several conditions, like different altitudes and camera shooting angles.
Furthermore, we report an in-depth experimental evaluation of several state-of-the-art object detectors, serving as baselines for further research on this topic.
Our analysis shows that detectors pre-trained on standard datasets of everyday objects exhibit moderate performance in localizing and recognizing people at sea in aerial images acquired at mid-high altitudes. The classification stage is the primary source of error for the best of the tested models, i.e., VarifocalNet, as about 82\% of the ground-truth persons were correctly detected but misclassified, thus suggesting that the same model fine-tuned on MOB data may have room for growth. To this end, as a future direction, we plan to extend our dataset with additional annotated data for the supervised training procedure, also considering synthetic images coming from virtual worlds. 

\bibliographystyle{splncs04}
\bibliography{biblio}

\end{document}